\documentclass{article} 
\usepackage{iclr2021_conference,times}

\usepackage{amsmath,amsfonts,bm}

\def\eqref#1{equation~\ref{#1}}

\def\1{\bm{1}}

\DeclareMathAlphabet{\mathsfit}{\encodingdefault}{\sfdefault}{m}{sl}
\SetMathAlphabet{\mathsfit}{bold}{\encodingdefault}{\sfdefault}{bx}{n}

\usepackage{hyperref}
\usepackage{url}

\usepackage{graphicx}

\iclrfinalcopy

\title{Maelstrom Networks}

\author{Matthew S Evanusa \\
University of Maryland, College Park \\
US Naval Research Laboratory, Washington DC \\
\texttt{matthew.evanusa@gmail.com}\\
\And
Cornelia Ferm\"{u}ller \\
University of Maryland, College Park \\
\And
Yiannis Aloimonos \\
University of Maryland, College Park 
}

\newcommand{\cfootnote}[1]{\let\thefootnote\relax\footnote{\centering #1}}

\iclrfinalcopy 
\begin{document}
\cfootnote{Distribution Statement A. Approved for public release: distribution is unlimited.}

\maketitle

\begin{abstract}
Artificial Neural Networks has struggled to devise a way to incorporate working memory into neural networks. While the ``long term'' memory can be seen as the learned weights, the working memory consists likely more of dynamical activity, that is missing from feed-forward models.   This leads to a weakness in current neural network models: they cannot actually process temporal data in time, without access to some kind of working memory.  Current state of the art models such as transformers tend to ``solve'' this by ignoring working memory entirely and simply process the sequence as an entire piece of data; however this means the network cannot process the sequence in an online fashion, and leads to an immense explosion in memory requirements.  In the decades prior, a separate track of research has followed recurrent neural networks that maintain a working memory via a dynamic state, although training these weights has proven difficult. Here, inspired by a combination of controls, reservoir computing, deep learning, and recurrent neural networks, we offer an alternative paradigm that combines the strength of recurrent networks, with the pattern matching capability of feed-forward neural networks, which we call the \textit{Maelstrom Networks} paradigm.  This paradigm leaves the recurrent component - the \textit{Maelstrom} - unlearned, and offloads the learning to a powerful feed-forward network.  This allows the network to leverage the strength of feed-forward training without unrolling the network, and allows for the memory to be implemented in new neuromorphic hardware. It endows a neural network with a sequential memory that takes advantage of the inductive bias that data is organized causally in the temporal domain, and imbues the network with a state that represents the agent's ``self'', moving through the environment.  This could also lead the way to continual learning, with the network modularized and ``'protected'' from overwrites that come with new data. In addition to aiding in solving these performance problems that plague current non-temporal deep networks, this also could finally lead towards endowing artificial networks with a sense of ``self''. 
\end{abstract}

\section{Introduction}

The ultimate goal of artificial intelligence  is to recreate the intelligence that biological agents have displayed to remarkable degree, extract the core components of intelligence, and reproduce this in an artificial agent, potentially amplifying this.  Since the early days of \textit{connectionist} networks, where we attempt to solve this intelligence problem through networks (or graphs) of simple artificial neural units, the goal has somewhat drifted away from a general artificial agent towards more specialized tasks, namely, pattern recognition.  Feed-forward neural networks, built off of the \textit{Perceptron} framework \citep{rosenblatt1961principles}, have excelled at pattern recognition, and have reoriented the entire field towards this mapping function.  This has culimated in the current generation of Large Language Models, which are at the core, meta-networks of Perceptrons trained using backpropagation \citep{vaswani2017attention}. What have we sacrificed in this focus on pattern recognition? This is the question that we motivate this work with.  We have, we argue, sacrificed the notion of the agent as a state, a sense of self, that persists across time, and the temporal correlations that accompany this. 

\subsection*{Data is Temporally Organized}

\begin{figure}[h]
\begin{center}
\label{fig:sequence}

\includegraphics[width=.8\textwidth]{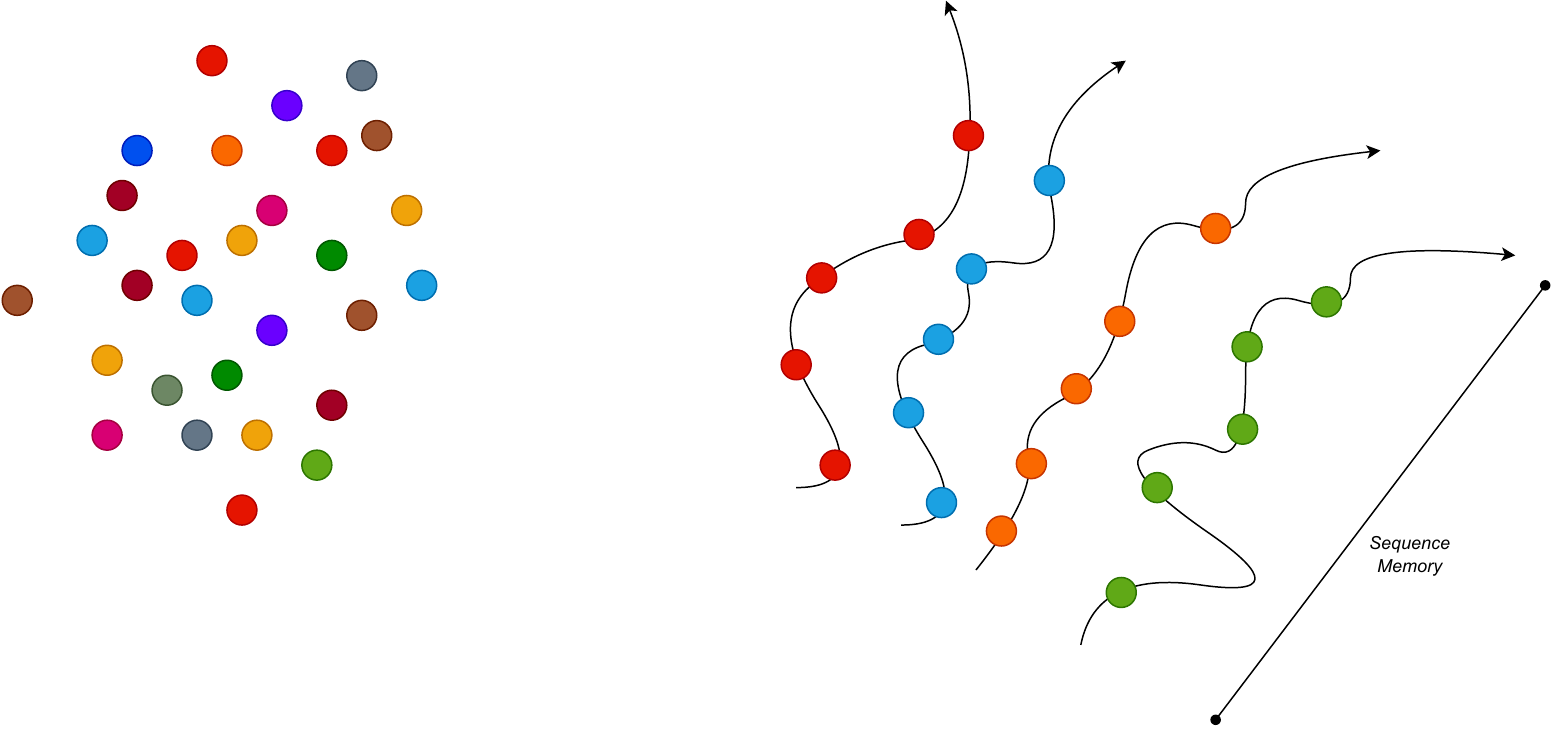}
\caption{As opposed to the current zeitgeist of machine learning, data in the real world follows inductive biases in how the data is structured not only in the spatial domain - for which we have taken heavy account - but also in the temporal one.  \textit{Left: } The current view of machine learning which sees data as independently and identically distributed (I.I.D) in time; it is the job of the network to learn the features in time that correspond to each data point.  This assumes the inductive bias of the spatial hierarchy of the data, but not the temporo-causal relationships of the data points along the same time thread. \textit{Right: } The new view of networks as also accounting for the temporal inductive bias of the data along the temporal dimension. The points are related to one another on the same temporal thread via the referential frame of \textit{action}: it is those actions which took a point from one location to another, as a result of the causal effect of the action. The ability of the network to recognize data along the same thread is what we refer to here as \textit{Sequence Memory}.}
\end{center}
\end{figure}

At the core of this issue is the current foundational viewpoint for deep learning that data in the universe is, in some sense, I.I.D, and has no underlying structure; it is up to the \textit{network} to learn this structure from random noise.  Creatures, however, evolved in the real world, with real physical limitations on the way that the system evolved to deal with, and survive in, the real world.  One of the set conditions that nature gives us, in addition to the 3-dimensional structure of the world, is the temporal nature of cause and effect; data later in time are temporally correlated with the earlier data. Figure \ref{fig:sequence} gives a graphical depiction of this dichotomy. The data that is aligned temporo-causally along the same strands we refer to as temporal sequences.  And the memory mechanisms in agents that is responsible for remembering points along the same strand we refer to as \textit{Sequence Memory}.  

The viewpoint of feed-forward connectionist networks, which is the current paradigm for deep learning, is that data must be assumed to be independently distributed - the I.I.D prior.  We assume that the data is ``given'' to us in a completely random, jumbled state, with no inductive prior on the structure of the data, and it is up to the network to learn the structure of the data.  This is what we expect from neural networks - the only inductive prior that we assume is that the data is structured \textit{hierarchically}, which is what leads to our ``deep'' learning structures of multi-layered networks.   We do not, however, flip this 90 degrees, and think about the structure of the data in time.  The data in the real world, however, \textit{is} structured: it follows clear causal relations between the data in time, which we can visualize as ``strands'' in time.  Each of the data that is causally linked sits on the same ``strand'', which may branch off from one another.  Figure \ref{fig:sequence} shows these two varying viewpoints of data in the world.  The story of deep learning has shown that the structure of the network topology is paramount: the learning rule (backpropagation) as well as the activation of the neurons has largely remained unchanged in the decades since deep learning came about, only the topology of the network, the structuring of the linear layers into convolution or attention layers, has changed.  This insight demonstrates that to effectively tackle networks in the real world, inductive biases must be given for the temporal domain and not just the spatial.  This inductive bias takes the form of \textit{Sequential Memory}: the ability of the agent to remember data that is causally linked accoring to the same temporal sequence strand.  This bias is implemented using modules of resonating, recurrently connected graphs of neurons which we call the \textit{Maelstrom}.

\subsubsection*{Brains Have State via Resonating Cell Assemblies}

How do creatures persist across time, and remember the memory of temporal sequences as they occur? Of course, statistical pattern recognition is a key component, but another component must be the system's ability to retain the information such that a pattern ``readout'' can occur.  This idea that the brain's mechanism for recognition is a combination of cell assemblies, which maintain activity, and a readout mechanism, which performs a mapping of activity to motor action and tasks, is well documented and is a current theory for brain organization \citep{buzsaki2010neural}.  According to the theories of Neural Assemblies by Donald Hebb \citep{hebb2005organization}, the activity persists in groups of neurons, called assemblies, that resonate their activity across time according to some stimulus.  It is this resonance, we argue, is the ``stable state'' that allows living creatures to persist across time.  A separate readout mechanism, then, performs mapping of these states, akin to a function learning a mapping of states of a dynamical system to outputs.   

\subsubsection*{Cell Assemblies are a Temporal Inductive Bias}

The key insight for the brain is that these assemblies \textit{model the temporal sequences by mirroring the cause-effect relationship of the input data, but in the recurrent patterns of the network}.  Thus, the same temporal patterns that occur in the real world, are ``echoed'' in the temporal patterns of the recurrently-linked cell assemblies.  This topographical structure of a recurrently connected graph structure that is modular and unhooked from the gradients of the sensory and motor areas are the temporal inductive bias in the brain, and in our proposed \textit{Maelstrom} network, that allows it to maintain a state and capture sequence memory. Whereas the spatial inductive bias is that the hierarchical nature of the features in the data are represented as feed-forward layers of the network, the temporal inductive bias is that the looped or cyclical structure of the recurrent components captures ``resonances'' which mirror the temporal-causal relationship of the data.  As the topology of the network in the brain is fixed, the specific resonances that occur will occur with the same temporo-causal patterns that occured in the real world, \textit{except} that it collects cause-effect from across time.

Interestingly, the story of deep learning has been one, we argue, of inductive biases and nothing more, in the form of network topologies, in the static space.  A deep multi-layered perceptron actually contains more parameters than the same layered transformer, however, the transformer performs remarkably better; this is generally understood to be because of the attention mechanism, which is just a special form of inductive bias on the structure of the linear layers; this structure biases the layers to learn representations that focus on attending to specific elements in a sequence.

\section{Limitations of Current Approaches}

Although deep neural networks have shown remarkable performance, their success masks some key missing elements if we are to achieve a thinking machine.  Although some posit that Large Language Models and this is an open debate in the literature, the current pinnacle of feed-forward networks, have ``reasoning'' capabilities, we would argue here that they rather are simply powerful pattern mappers, and that many tasks we acribed to reasoning in the past are actually just complex mapping tasks.  This is a longer debate outside of the scope of this paper, but for the purposes of this work, we argue they cannot be true reasoners in the real world as they are not embodied. And to be embodied, they are missing one critical components: namely a memory of the past activations via a \textit{state}.

\subsection{Feed-Forward Networks Lack Sequence Memory}

Memory, as a concept, is tricky to pin down, although there have been some interesting definitions of memory that bridge the biological with the computational \citep{zlotnik2019memory}.  Here, we take \textit{memory} in the context of connectionist neural networks to mean: \textit{a mechanism by which the neural network maintains a state of itself, and contains information about the past from a temporal sequence}.  This is what we refer to as described earlier, as \textit{Sequence Memory}.  The idea of sequence memory is intricately tied to the notion of time: to process time, an agent must record, in some state variable, some information of timesteps prior.  As articulated in earlier sections, the idea of a recurrent network, unhooked from the gradient, gives a temporal inductive bias to the processing of the data.  This accompanies with it the important notion that \textit{all} data we receieve, as active agents \citep{friston2016active, aloimonos1988active}, is in the temporal domain.  In our own brains, we do not have access to a ``RAM-like'' memory that computer systems have (as our entire system is the connectionist neural network), the neural network itself must keep a memory of its own activations, to serve as the state variable for temporal processing.   This ``state'' of the neural network is then updated with new information, in a recursive fashion.  This kind of architecture is mirrored in Recurrent Neural Networks (RNNs), as described below, and serves as an important foundation for our proposed ideas here. 

Of course, incorporating a ``state'' that updates with the model introduces a host of challenges.  It is not hard to see why the current state of the art neural networks, built on Transformers \citep{vaswani2017attention} which then became Large Language Models \citep{radford2019language}, for the most part eschewed the idea of recurrence (although some hybrid work has arose in recent years attempting to fuse them \citep{hutchins2022block}).

\subsection{Memory Serves in the Service of Embodiment}

To be an artificial embodied intelligence \citep{chrisley2003embodied,duan2022survey}  entails placing the learning system in a chassis that can take actions in an environment. This is an active process \citep{aloimonos1988active,friston2016active}, with the agent contained in a control loop within the environment. And to realize these actions in an environment which is moving in time, the agent must have access to a running state of the previous activity.  To not have this state would amount to the agent running with ``blindfolds'' on around their perception completely, only seeing the current timestep and making a decision, then moving onto the next sensory perception.  This realization amounts to reducing agents to mere pattern recognizers - thus the idea of having a memory in a strong sense elevates agents above merely pattern matching entities.  It has also been argued that the entire purpose of memory, rather than serving as some RAM-like appendage that stores abstract values for later use, completely serves embodiment \citep{glenberg1997memory}.

\section{Prior Work on Connectionist Memory}

\begin{figure}[ht!]

\centering
\includegraphics[width=\textwidth]{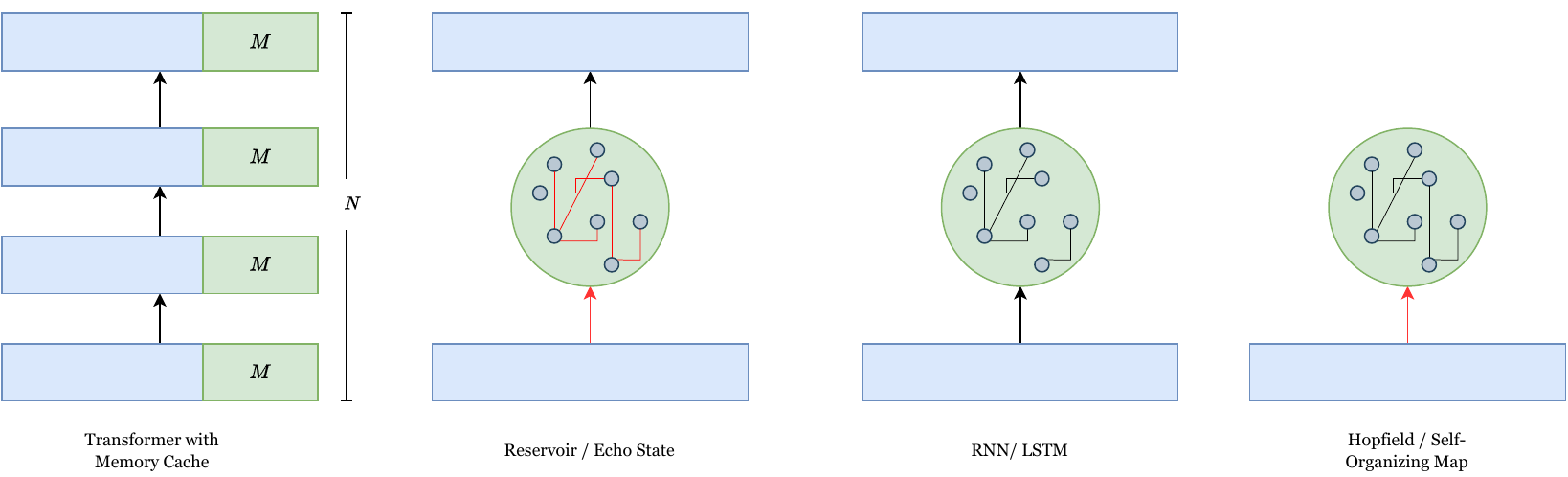}

\caption{Comparing the various prior approaches to dealing with temporal sequential memory (or lack thereof). Blue rectangles indicate a feed-forward network layer (no memory), while green represent a recurrently connected one or memory that is accessible through timesteps.  Black arrows represent a learnable weight, whereas red indicates unlearned, where the gradients do not pass backwards.  \textit{ From left to right}: Transformers or CNNs \citep{lecun1998gradient,devlin2018bert,vaswani2017attention}  are in their vanilla state purely feed-forward networks, and are not represented here because they do not contain any form of sequence memory.  Only newer variants of Transformers such as Transformer-XL, which contain a ``cache'' that is accessible to later timesteps, have what we would consider sequence memory.  However, these are not ``true'' sequence memory as they do not solve the continual learning problem, due to the fact that they are still non-modular \citep{hadsell2020embracing} (i.e., the memory component is still ``hooked'' to the motor networks) and thus any new gradients would overwrite previous timesteps.  Reservoir networks (echo state networks or liquid state machines) have a recurrent component with unlearned recurrent weights and input, but a learned readout that can be a feed-forward network \citep{evanusa2023t}.   RNNs and LSTMs have a recurrent component where every connection is learnable. Both LSTMs and Reservoirs can have multiple recurrent ``layers'', connected in a hierarchical structure \citep{gallicchio2017deep}.  Lastly, Hopfield \citep{hopfield2007hopfield} and Self Organizing Maps  \citep{kohonen1990self}  are recurrent components without a learned readout, where the recurrent weights are trained using an unsupervised self organizing rule.}
\label{fig::maelstrom}
\end{figure}

\subsection{Memory in Connectionist Networks}

As mentioned earlier, the way that researchers think about neural networks \textit{in silico}, versus how neural networks work \textit{in vivo}, is fundamentally different not only from the algorithmic perspective, but also with respect to how memory is stored.  In computers, we simply store our neural network code in RAM or disk storage, which is a collection of buckets that can store any arbitrary numerical values (without respect to any task).  All of the functions that require remembering (such as loading code, the weights, the dataset) are stored in this disk memory.  In contrast, in the brain there is no arbitrary storage separate from the neural network - the storage \textit{is} the neural network.  This leads to some more complex and dynamical architectures to store values in ``working memory'', versus the ``long term memory'', which corresponds to the disk memory.  Working memory \textit{in silico} corresponds more closely to RAM in that a consistent voltage must be applied, although this memory is stable and unchanging, whereas dynamical attractors in the brain are constantly in flux.  Regardless, in connectionist networks, this memory must be stored as a \textit{state} of the network, or an abstract vector that updates in time as the network progresses through inputs.  As the brain does not actually have this vector which itself would be stored in disk, the \textit{state} of the brain is really the current snapshot of the voltages and accumulation of neurotransmitters for each neuron terminal and body.  To read the state, the brain has only access to the action potentials or ``spikes'' that are emitted to sent information between neurons (barring any unforseen tricks by the glia).  Thus, to truly implement this state in a neuromorphic way, the value of the state is simply the output activity being sent back to itself, creating a ``self loop''.  While in code it is possible to maintain in the disk the states of the ``transmitters'' of the neurons, this idea of recurrent connections is still critical as loops provide recurrent computation and allow for the memory to reverberate throughout the network. Thus, the network persists its state (or memory) of the temporal signal by reverberating or bouncing a reflection of the input inside itself to maintain a persistent activity. 

\subsection{Hopfield Networks}

A result of Donald Hebb's seminal work on neural population and organization \citep{hebb2005organization}, Hopfield derived a network that views the network as an energy minimization problem, where the activity that bounces around and settles into \textit{attractor states}, and the learned patterns correspond to energy minima in this landscape.  It has been shown that using the Hebbian learning rule mathematically equates to finding, for a given data sample, the energy minimum attractor state associated with that network architecture.  This has the benefit of not needing labeled samples, is completely self-organizing, and is biologically relevant.  The issues hampering it have been a lack of ability to extract latent codes - features -  from large amount of samples, and the somewhat strict limits on memory that result from this.  However, new work has shown that via gating mechanisms \citep{hochreiter1997long, davis2022neurolisp} the memory can be increased. The gradient-based LSTMs also suffer the same problem as in the earlier section about the weights not being sensitive to the particular timestep. While in their basic form Hopfield networks have failed to match deep learning, newer architectures that incorporate continuous values and attention show correspondence between Hopfield learning and attention layers in Transformers, and are a promising direction for future RNN research \citep{ramsauer2020hopfield}.

\subsection{Gradient-Based RNNs}

A simple approach to training a recurrently connected network is to treat it like a feed-forward network, and treat the time dimension as if the network were actually multiple layers deep. For a single-layered RNN,  the network is run in time for a set number of $t$ timesteps, and then the network is ``unrolled'' to have $t$ layers corresponding to the $t$ timesteps.  The gradient is learned and then the updates are all applied to the same weight vector.  This training mechanism is known as ``Backpropagation Through Time'' (BPTT), as the network through time is unrolled and treated as a deep neural network. It is generally accepted that BPTT is highly biologically unrealistic as the brain cannot unroll itself, but several approximations have been developed in recent years that try to argue for a mechanism in the brain that produces similar effects \citep{cheng2023replay, manneschi2020alternative}.

The issue, still, is that the memory and the feature vector are not truly topologically separated: the ``memory'' vector (represented ostensibly by the cell state) is still trained and \textit{driven} by the same signal as the error for the feature vector.  The only difference is that the cell state `prefers' to stay unchanged for longer periods, allowing for long-term dependency learning; this is not a true philosophical ``separation'' of the memory and the feature vector.  Thus, it is not surprising that the Gated Recurrent Unit or GRU \citep{DBLP:journals/corr/ChoMGBSB14}, which fuses the cell state and the hidden state - a process that if the memory were truly separated and the system depended on it, would destroy the function of the system - actually tends to perform the same or sometimes even better, with less overhead, and is the preferred RNN of choice in modern times when RNNs are used.  Of course, RNN usage has been in recent times completely eclipsed by feed-forward networks - in particular Transformer models \cite{vaswani2017attention}.

\subsection{Reservoir Computing}

As a response to the requirement of stable attractor states from Hopfield Networks, Reservoir Computing \citep{jaeger2001echo, maass2011liquid} was developed to convert the problem from an attractor-based one to a mapping based one.  Critically, it converts the problem of having the network settle, to having the network ``observe'' the dynamical state of the recurrent state (now called the \textit{reservoir states} - but effectively the same as a hidden state of an RNN).  The insight is that we need not propagate the error through the recurrent component.  The component will produce some activity as a result of its initialization, and if that initialization is random, or is a sufficient basis set to cover the possible features, a sufficiently powerful readout is capable of mapping that activity to any predictive value.  Here, it is also important to note that we need not learn, in the recurrent component, \textit{what} the memory activity corresponds to; this is the task of the readout mechanism.  The recurrent component simply is in charge of bouncing and persisting activity, just long enough for the readout to make a determination.  This new paradigm fundamentally shifts around the requirements of a memory mechanism, and makes an important founding that we use for this work as well: that the memory and predicitve processing components can potentially be split into two separate processes.  While the memory necessarily must be tuned to help in the service of prediction as in \citep{glenberg1997memory}, it is really the combination of this readout mechanism and the memory that constitutes the effects of memory.  We take this idea of separating the memory and computation components, along with the idea that we need not propagate the gradients through the recurrent component, as the job of the ``observer'' is simply to perform a mapping. 

One of the key insights of reservoir computing that will be extracted is that the memory unit is topologically separated, \textit{unhooked}, from the gradient learning processs of the \textit{readout}, which is the network assigned to do the functional mapping from states to actions (labels, or values). We take this core idea from reservoir computing, and expand it outward in the \textit{Maelstrom} paradigm into a much larger proposed theoretical structure, without the limitations placed by reservoir computing.



\begin{figure}[ht!]

\centering
\includegraphics[width=\textwidth]{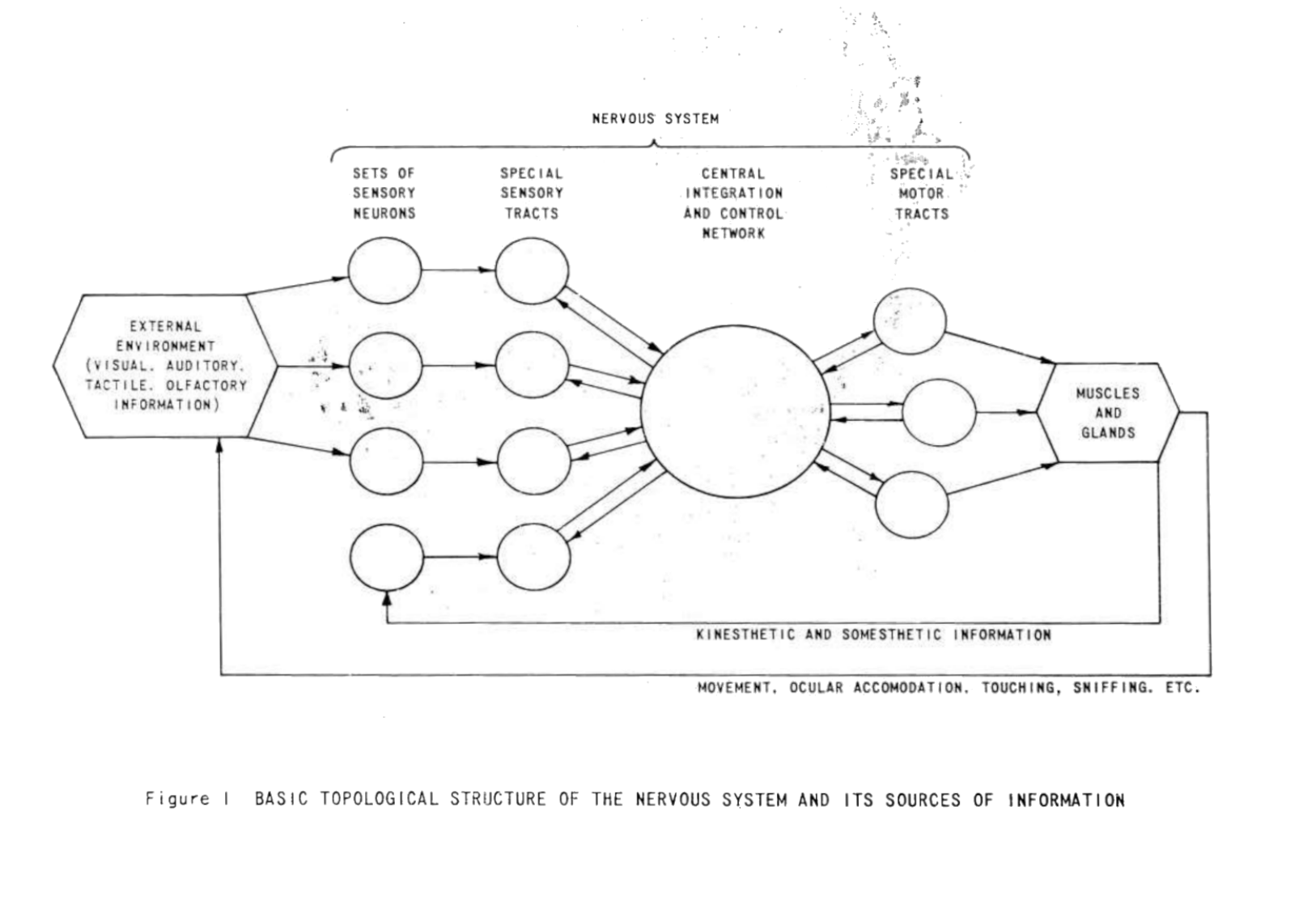}

\caption{The topological organization of the nervous system as proposed by Frank Rosenblatt in 1961, taken from \cite{rosenblatt1961principles}. This bears a strong resemblance to the Maelstrom Paradigm: The sensory tracts and memory represent the input function, the motor tracts represent the output function, and the integration network is the Maelstrom.  The only difference, and the only thing lacking from Rosenblatt's account in our opinion, is the notion of sequence memory.  The Maelstrom can be seen as an implementation of Rosenblatt's ideas, in combination with deep learning as well as notions of state and sequence memory.    }
\label{fig:rosenblatt}
\end{figure}

\section{Maelstrom Networks}

To solve the issue of sequence memory and to endow a neural network with its capabilities, we present here \textit{Maelstrom Networks}. From control theory, we take the idea of a state space model that incorporates a memory state, an input function, and an output function.  From reservoir computing we incorporate the notion that the temporal memory must be a process topologically separated from the readout mechanism, in terms of learning via the gradients. And we use feed-forward networks for their preferred role: to map inputs to outputs, and not to learn how to store things in memory.   In Fig. \ref{fig::maelstrom} we show the schematic overview of the proposed approach. From an implementation standpoint, the input and output functions are easily parameterized by deep neural networks - blocks were shown here for visual simplicity. The key element which gives the network its name is the \textit{Maelstrom}, a recurrently connected component which can be implemented as a reservoir - but not exclusive to - , that takes input from the input function (an input neural network), bounces this activity around, and passes this to an output function.  we say ``passes to'', but in reality what is happening is, it is more akin to the output function reading out the maelstrom; you one can visualize this as the output function being a pair of eyes that observe the chaotic activity of the maelstrom.  Because the network is untrained and recurrently connected, it is a potentially chaotic system - a storm that gives the maelstrom its name.  It is the job of the input to ``control'' the activity of this storm and keep it within acceptable bounds.  

Critically as well, the maelstrom possesses ``top-down'' feedback onto the input function, which allows the feed-forward input function a control loop over its own activity.  This ``closing the loop'' on the malestrom is what gives a meta-loop between the maelstrom and the network structure as a whole.  These feedback connections are ubiquitous in the primate cortex \citep{zagha2020shaping} and it is this loop that feeds back into the input which, in addition to the loop within the maelstrom, creates the ``state'' that we percieve as our continuous self.  The inclusion of this loop is intended to, for the first time, imbue this sense of ``self'' with deep learning. 

The stimulus from the outside world is passed first through the input function - parameterized by a deep \textit{feed forward} network.  This does exactly what neural networks are good at: function mapping; the job of the input network is to map  The activity is passed from the input controller to an interface (another neural network) that may or may not connect around the maelstrom as a skip connection (dotted line - Fig. \ref{fig::maelstrom}). The job of The key element is that the gradient does not flow through the maelstrom - it simply agglomerates activity from the neural network surrounding it.  And critically, the output function \textit{sees the maelstrom as if it were the input}, as the maelstrom does not have access to the gradients of the output.  This is akin to the idea of ``neural ensemble readout mechanism'' as summarized in \citep{buzsaki2010neural}. The maelstrom accrues a memory of activations, which can be guided by the input neural network as well, since those weights are also learned.   The term  \textit{Maelstrom} represents the fact that this memory is a chaotic storm of activity that we can see from the outside, but cannot access the internals of.  It's important to stress that this does not force a specific setup for an architecture, but rather a general class of architectures that need just follow this general setup of the gradient flows.  The critical feature for the maelstrom network is a disjointed memory component where gradients cannot flow through, but that give the output function access as an input. This bears a strong resemblance to Rosenblatt's early work on the topological structure of the nervous system, as seen in Fig \ref{fig:rosenblatt}. 

The Maelstrom Network's only requirements are the structure of the input, and output, and recurrently-connected maelstrom that feeds back into the input of the input network.  How one implements these is up to the user, however, it is clear that a deep feed-forward neural network is primely situated to serve as the input and output functions, and a recurrent network in some capacity is primed to serve as the maelstrom.  How complicated these are, and how the gradient is passed around the maelstrom, is still an open area of research. In a more philosophical sense, we can think of the cell assemblies as solving the issue of sequential memory by ``mirroring'' the activity out of the outside world, but in the internal model of the agent. 

It is our opinion that the best implementations for the maelstrom, however, build on the work that demonstrate that the brain does behave like a near-chaotic system; the ability of the output function to map chaotic network states to actions is a key capability that deep learning offers us, in parameterizing with them.

\begin{figure}[ht!]

\centering
\includegraphics[width=.8\textwidth]{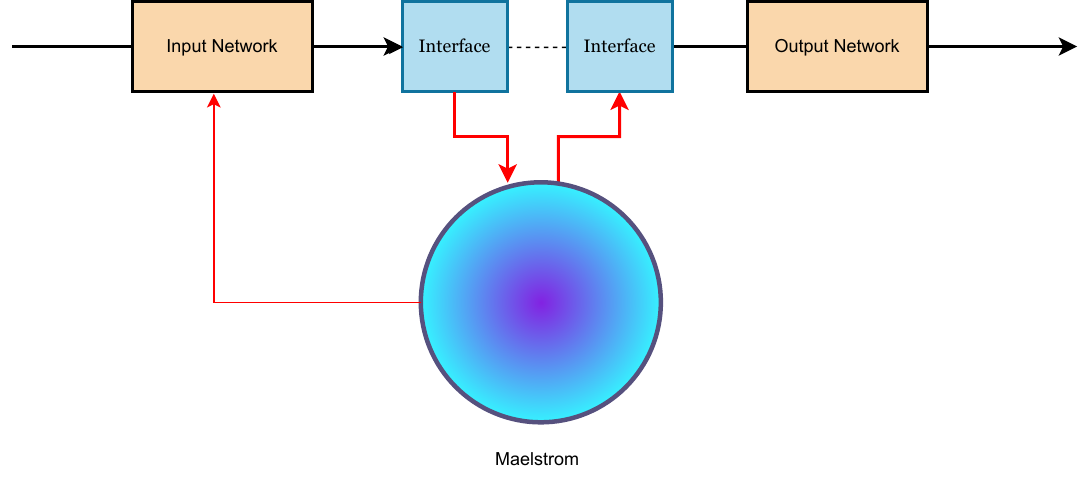}

\caption{The Maelstrom Network paradigm.  Arrows indicate The input is passed through an \textit{input network}, a feed-forward neural network which maps input patterns to control actions on the maelstrom.  This is passed then to an interface, which serves as the hub for talking to and from the maelstrom.  The interface passes to the \textit{maelstrom}, a recurrently connected state space that collects and conglomerates actions from the controllerl. The maelstrom bounces and maintains a state of the previous input.  As the maelstrom is recurrent and unlearned, or unhooked, from the gradient of the output, it exhibits chaotic behavior - it is the job of the input network to control this activity. The interface then reads the maelstrom and passes this to an output function, which then produces an output.  For neural network approaches, the input function, output function, and interface are all multilayered neural networks. Critically, when learning, the gradients cannot flow through the maelstrom; this entails the exhaustive unrolling of the network to compute accurate gradients, and its highly biologically unlikely. In contrast, the maelstrom does not need to unroll, which makes it more attractive as well for a biological model that is able to account for the randomness of connections while also preserving computational power. Black lines indicate connections where error can back-propagate and induce learning, red indicate connections that do not allow error to back-propagate.  This also allows for a ``skip'' connection  between the interface components to assist in gradient propagation (dotted line).  }
\label{fig::maelstrom}
\end{figure}

\subsection*{Relationship to Control Theory}

The structure of the network bears a strong resemblance to two well-known results from control theory.  The setup of the connection closely resembles a \textit{state space model}, with the input function taking the place of the $B$ matrix, the Maelstrom representing the state variables $\textbf{x}$,  and the output function r matching the $C$ output matrix. The maelstrom network can be seen as a case of \textit{nonlinear state space models}.  

In this setup, the input and output networks are trained using the MIT Learning rule \citep{mareels1987revisiting}, where the system is aiming to control via samples from the Maelstrom (which it has no control over), and uses gradient-descent to lower its error over time.  One way to view the Malestrom setup is a fusion of the MIT learning rule, a recurrently updated state vector, and deep neural networks. 

\subsection*{Relationship to Reservoir Computing}

As the idea of the malestrom was born about of reservoir computing \citep{evanusa2022deep}, it is natural that it would be connected.  In \citep{evanusa2022deep, evanusa2023t}, what can be thought of as a ``partial maelstrom network'' had been introduced; this included ideas of the maelstrom - the recurrent component - and a readout (or, output network) for a task.  It did not, however, include the missing component, which was the sensory cortices.  With the input network attached, the network is ``complete'' in its development phase.  However, this does not mean that a simple randomly connected reservoir is the only possible maelstrom - it was used for simplicity and proof of concept, but in our proposal here, we envision any complex recurrently connected state machine that is topologically disconnected from the readout mechanism.  

\section{Advantages of Maelstrom Networks}

As the Maelstrom network paradigm is inspired by -  and can be seen as an evolution of -  the reservoir computing paradigm, it inherits many of the benefits. It also collects new advantages, as it incorporates deep learning into the mixture in ways that reservoir computing had not. 

\subsection*{Provides a Model for Theoretical and Systems Neuroscience}

\begin{figure}[ht!]

\centering
\includegraphics[width=.8\textwidth]{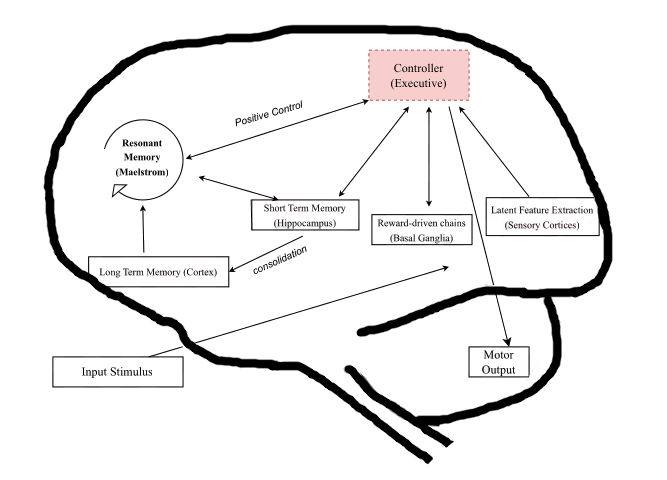}

\caption{The proposed relationship of the Malestrom paradigm with functional modules in the human brain.  One of the main thrusts of the Maelstrom paradigm was to create a model that both provides advancements to artificial intelligence, and to human brain understanding, simultaneously; this is the ideal goal of artificial intelligence research.  The stimulus enters the brain through the sensory cortices, such as the visual and auditory cortex.  These cortices contain top-down feedback control (which is exemplified by the feedback from the maelstrom to input controller), but are seen as functionally feed-forward (i.e., the gradients do not pass recurrently).  These sensory cortices map to the input network of the Maelstrom.  These features are passed to the \textit{executive} modules, which are a large open question in neuroscience as to their location, but we suggest they exist likely in the hub regions that control, recieve from, and regulate multiple regions.  In the Maelstrom Paradigm,  the executive is rolled into the input controller network, however we envision future work as a separate executive module distinct from the sensory network.  This network then sends data to the Maelstrom, where it bounces around recurrently in a ``storm'' of chaotic activity; this is in our view located as a mix of the prefrontal cortex as well as the hippocampus. Also left for future work is consolidatio of memory, or learning of memory features, in the maelstrom, as it is left completely untrained in the current simpler iteration. Lastly, the output is sent (either from basal ganglia-learned actions, explicit control, or reflexes), through the executive, to the cerebellum where it learns the correct weights for mapping these to motor actions. The positive feedback control from the maelstrom back to the controller is represented in the maelstrom network via the feedback connection from the maelstrom to the input network, this creates a larger meta-loop within the system at a level above the loops within the maelstrom, and contributes as well to the phenomenon of ``self'' of the system.}
\label{fig:brain}
\end{figure}

As the Maelstorm paradigm was inspired by reverse-engineering the brain, it is possible to create a ``two-way'' highway between this proposed model and the brain function, where the brain informs AI research, and the AI research informs neuroscience.  Figure \ref{fig:brain} demonstrates our general view of the brain's interactions, and how this can map to the maelstrom paradigm. Each module of the Maelstrom Network maps either to a single area of the brain, or a group of areas.  This modular structure of the brain, also echoed in \citep{hadsell2020embracing}, is essential to performing the complex tasks, in a way that does \textit{not} overwrite weights, which occurs when we train networks using the I.I.D assumption on temporal data. The goal here is to create a system that both advances the field of A.I. \textit{and} advances the field of neuroscience simultaneously. 
In addition, we propose that it is through these multiple-hierarchy loops at varying scales, i.e. within the maelstrom and then within the feedback loop from the maelstrom and the controller, which provides the ``state'' that serves as the continual sense of self that agents phenomenologically percieve.  This could potentially link systems neuroscience structures with concepts of consciousness.

\subsection*{Encapsulates the Embodied Notion of Memory}

It is clear that while computer memory relies on memory as an abstract storage, memory in the brain likely evolved for a specific goal, which was to enable the survival of the agent.  This means that memory serves at the behest of embodiment, and not simply as an abstract bin that any information can be stored into, as argued in \citet{glenberg1997memory}. This is also reflected here in the Maelstrom paradigm, in the combination of the memory and the interface.  The maelstrom's memory itself is a meaningless string of information, that \textit{only} has meaning with respect to a given readout - the readout which maps the information to a specific prediction, or task.  This is one potential solution to the issue of \textit{grounding} of the meaning of activity - while the activity itself is potentially meaningless, it is given embodied meaning by virtue of it being tied to a specific readout.  This is also the view espoused in the Neural Assembly theory of \citet{buzsaki2010neural}.

\subsection*{Enables powerful training of feed-forward networks while combining Memory}

It is abundantly clear that feed-forward neural networks work best with backpropagation, as Transformers \citep{vaswani2017attention} have completely overshadowed recurrent neural networks for sequential processing, even though Transformers contain no memory themselves in their vanilla form (see \citet{dai2019transformer} for a version with a cached memory).  While it is true that the Maelstrom removes the ability to attend to far-back sequence elements, this \textit{is necessarily what must happen} if we want to process sequences in real-time, as described below - the only way to attend a weight for each past element is to give the network access to the past elements, which would in turn reduce the model back to a Transformer.  In fact, combining a recurrent memory with attention actually \textit{predates} Transformers themselves, as in \citep{chorowski2015attention}, and the main innovations of the Transformer were simply \textit{removing} the LSTM \citep{hochreiter1997long} component (hence the ``All you need'' of the title) and adding the multi-headed features.  In this case as in \citet{chorowski2015attention}, the readout attention mechanism still needs access to the entire length of the sequence, even though it has a memory, because of these limitations being placed by the readout attention, and the fact that it must pass gradients through the LSTM to learn the weights. The Maelstrom paradigm also allows for additions to learning or attending to the Maelstrom - its requirement is simply that the gradients cannot flow back through it.  Naturally, adding in a memory necessarily means aggregating the past in some way into a compact latent representation, which does lose information in the process - there is no avoiding this.  The fact that Maelstrom Networks are a general principle also means there is room to incorporate new elements in the system.  For example, to enable even more associative memory, the Maelstrom can be trained using the Self Organizing Map \citep{kohonen1990self} procedure to agglomerate similar representations.

\subsection*{Easily Incorporates into Existing Feed-Forward Network Structures}

Rrecent work as in \citep{hutchins2022block,wu2020memformer} have begun to investigate using cached memory representations to alleviate the computational burden that Transformers have when dealing with long sequences.  Howeveer, these architectures requires highly specific structures for the memory to incorporate them in, and some are highly tuned towards language models and not general temporal data.  The Maelstrom paradigm, in constrast, requires no strict guidance on what the memory must look like.  The key insight is that \textit{the memory representation passing through the maelstrom is fed into the readout, in such a way that to the readout, it appears to be input stimulus.}  This reflects a nested hierarchical view that \citet{friston2016active}  touches on; this notion that to sub-regions of the network, input coming in appears the same as the stimlulus does to the input layers of the network. This general paradigm, we believe, applies to all temporal data. 

\subsection*{Allows for real-time learning and inference of temporal data}

One of the major strengths of Transformers, their learning ability through feed-forward networks, is also one of their greatest weaknesses, as they must take the entire length of the sequence to process the sequence. While it is possible to train a Transformer to map all possible sub-sequences in a feed-forward manner, this becomes computationally infeasible very quickly, as the number of sub-sequences grows as $N^2$ with the number of sequences. 

Adding this Maelstrom, however, that is completely unhooked from the gradient learning, turns the problem of learning sequences into a controls problem, where the input at each timestep is simply a feed-forward learning of the current memory state vector, and its mapping to a prediction.  This means that the entire length of the sequence is \textit{no longer needed}, which means that the sequence can be learned in an online fashion, \textit{and} that inference can be performed in an on-line fashion.  This is critical for any use-case where neural networks need to be running in time in the real world - which, by most measures, would include most real-world applications. 

\subsection*{Allows for neuromorphic hardware implementations}

Lastly, as the maelstrom does not require gradients to backpropagate through the recurrent component, it is extremely amenable to newer neuromorphic and bio-inspired hardware implementatoins, such as through the Intel Loihi \citep{evanusa2019event}, or Memristor technologies \citep{thomas2013memristor}, or even FPGA implementations.  These technologies eschew with the traditional CPU chip processors and instead solely work by modeling \textit{just} the neuron activations.  This reduction of the function of the chip severely limits the kind of processing it can do (for normal computer programs), but supercharges the use-cases when it runs neural networks.  As the memristor and loihi chips in particular have specialized hardware, performing Backpropagation Through Time or other complex temporal operations are burdensome; running a maelstrom, however, that requires no internal updating and simply runs as an autonomous dynamical system, is much more doable.  The fact that the Maelstrom is disjoint from the feed-forward component means that the feed-forward component can be implemented separately in GPU hardware, and read from the neuromorphic Maelstrom when needed.  This would fully utilize the power saving capabilities of neurmorphic hardware (as the recurrent component is computationally expensive with BPTT), in a time when it is becoming abundantly clear that the energy usage for current deep learning is unsustainable.

\subsection*{Allows for Online or Continual Learning}

As the sequence memory, encapsulated in the maelstrom, is unhooked from the gradient learning process, this can allow future instantiations to deal with the \textit{continual learning} problem.  The continual learning problem is defined as the ability of a neural network to train indefinitely while it continues to perform testing inference.  This is impossible with current deep learning networks due to the fact that the gradients from new samples completely overwrite the old weights with no regard to their importance.  For example, a network trained on five image classes, but then continued to train on 5 new classes without access to the old classes, will completely forget the first data.  It has been proposed in \citep{hadsell2020embracing} that a modular architecture, for example the Maelstrom paradigm, can aid in the continual learning problem, by unhooking the gradients, and in addition by removing the I.I.D assumption that underlies the current setup. The unhooked Maelstrom will allow for the network to not overfit to a given task along the sequence, while also not overwriting later sequence elements.  

\section{Conclusion}

In this work, we propose \textit{Maelstrom Networks}, a novel modular neural network architecture that unhooks a sequential memory, representing the prefrontal cortex and hippocampus, from the gradient pass of two feed forward networks used to control the memory.  The input network acts as the sensory cortex to the system, inputting the correct features while maintaining balance of the maelstrom, and the readout takes outputs from the maelstrom and applies them to an action or task, similar to the motor cortex.  We propose that this work is the natural evolution of work done in the early 1960s, which laid out a larger modular structure for the nervous system but left out the critical issue of sequence memory.  We connect this work to control theory and note that the maelstrom is a nonlinear state space model, trained using the MIT control rule; this also relates the brain to notions of control as well, as we map each region of the brain to components of the maelstrom as well.  It is our hope this helps usher in a new era of neural networks that specifically focus on sequence memory, not as an engineering trick to aid in performance, but as a fundamental property of networks.



\bibliography{iclr2021_conference}
\bibliographystyle{iclr2021_conference}


\end{document}